# AI shares emotion with humans across languages and cultures

Xiuwen Wu[1,2], Hao Wang[3,4], Zhiang Yan[2,4], Xiaohan Tang[2], Pengfei Xu[5], Wai-Ting Siok[6], Ping Li[6,7], Jia-Hong Gao[2*], Bingjiang Lyu[8*], Lang Qin[2*]

1. University of Science and Technology of China, Hefei, China
2. Peking University, Beijing, China
3. Tsinghua University, Beijing, China
4. 01.AI, Beijing, China
5. Beijing Normal University, Beijing, China
6. The Hong Kong Polytechnic University, Hong Kong SAR, China.
7. The PolyU-Hangzhou Technology and Innovation Research Institute, Hangzhou, China.
8. Changping Laboratory, Beijing, China

*co-corresponding authors

## Abstract

Effective and safe human–machine collaboration requires the regulated and meaningful exchange of emotions between humans and artificial intelligence (AI). Current AI systems based on large language models (LLMs) can provide feedback that makes people feel heard. Yet it remains unclear whether LLMs represent emotion in language as humans do, or whether—and how—the emotional tone of their output can be controlled. We assess human-AI emotional alignment across linguistic-cultural groups and model-families, using interpretable LLM features translated from concept-sets for over twenty nuanced emotion categories (including six basic emotions). Our analyses reveal that LLM-derived emotion spaces are structurally congruent with human perception, underpinned by the fundamental affective dimensions of valence and arousal. Furthermore, these emotion-related features also accurately predict large-scale behavioural data on word ratings along these two core dimensions, reflecting both universal and language-specific patterns. Finally, by leveraging steering vectors derived solely from human-centric emotion concepts, we show that model expressions can be stably and naturally modulated across distinct emotion categories, which provides causal evidence that human emotion concepts can be used to systematically induce LLMs to produce corresponding affective states when conveying content. These findings suggest AI not only shares emotional representations with humans but its affective outputs can be precisely guided using psychologically grounded emotion concepts.

## Introduction

Emotion is a cornerstone of human intelligence, and language is one of its most important carriers[1,2]. An old friend can easily discern the sadness beneath a seemingly light-hearted remark, and the nuanced emotions expressed in literary writings transcend time and space to move readers. Endowing AI with emotional intelligence comparable to that of humans is not only a long-standing open question, but also an aspirational vision. With the rise of large language models (LLMs)[3,4], AI is becoming an integral part of our daily lives[5], fostering social systems based on human-machine coexistence[6,7], and the emotional alignment between humans and AI becomes paramount. An essence of empathy refers to the sharing of emotions between individuals, promoting mutual understanding and trust[8]. Thus, effective and reliable human-machine collaboration necessitates the regulated and meaningful sharing of emotions between humans and AI systems. Without a degree of emotional congruence—where AI can appropriately interpret human emotion and generate emotionally resonant responses, and humans can understand the basis of AI's emotional expressions—our collaborative endeavours with AI may be hindered by a lack of trust and togetherness[9-11]. More importantly, as generative AI expands into application domains that require nuanced emotional understanding, managing the emotional states of LLMs can help ensure safer and more ethical human-AI interactions[12]. However, it still remains unknown whether LLMs represent emotion in language as humans do, and whether—and how—the emotional tone of their output can possibly be regulated.

Efforts have been made to examine emotional processing in both humans and machines, leading to reciprocal inspirations[13-15] and even to enabling AI to emulate emotions. Still, investigating and achieving human–LLM emotional alignment is challenging. First, the foremost obstacle arises from the inherent complexity and subtlety of human emotions, with cultural and linguistic differences imposing additional barriers. Our emotions are not static, discrete categories but dynamic, nuanced experiences that vary in intensity[16], significantly shaped by cultural contexts. Classical emotion theories identified basic categories[17,18] and dimensional frameworks like valence-arousal[19,20], while computational semantic space approaches sought to capture more nuanced emotional variations, identifying broader and more comprehensive categorical sets with fuzzy boundaries[21,22]. For example, "happiness" can range from mild contentment to intense euphoria, rather than being a single uniform category; at the same time, culture-specific emotional states may not easily be translated or understood across languages, with countless examples such as Japanese "物の哀れ[1]" or German "Waldeinsamkeit[2]". Second, the challenge of reaching robust alignment between LLMs and human emotion also stems from the nature of data typically used in affective computing. Affective computing has synthesized cognitive research with advanced computational methods, significantly advancing machine recognition and interpretation of human emotions[23,24]. Common emotion datasets often operationalize affect through discrete categorical labels (e.g., Ekman's basic six emotions) or simplified dimensional scales (e.g., positive/negative sentiment). Such coarse-grained representations are often insufficient to capture the nuanced, continuous, and high-dimensional characteristics inherent in the human affective experience[21,22], which constrains the development of internal models that fully correspond to the complex and dynamic nature of human emotional phenomena.

Given that semantic representations[25,26] and contextual information contain rich emotional content[27,28], it is plausible that LLM-based AI systems are capable of capturing the emotions

[1] the pathos of things
[2] the feeling of solitude in the woods

embedded in language even if they do not possess emotions of their own. Recent advances have highlighted the substantial potential of LLMs for sophisticated emotional processing[29]. In practice, emerging evidence indicates LLMs can integrate contextual information to interpret nuanced emotional content[30,31], and such ability has been compellingly demonstrated in text-based interactions in different languages[32,33]. Latest research reported that LLMs can acquire emotional knowledge in the absence of sensory-motor grounding, underscoring the role of language-derived conceptual representations in emotion inference[34]. Despite these advances, foundational limitations remain unresolved. First, due to a lack of interpretability, the relationship between LLMs' internal emotion-specific computational features and human emotional perception is still poorly understood, posing a critical barrier to achieving deeper human-AI emotional alignment. Even though attributes of human emotion concepts were found to be encoded by distinct populations of artificial neurons within LLMs[34], individual neurons in LLMs—their natural computational units—are typically polysemantic, responding to diverse and seemingly unrelated inputs, making them challenging for direct human interpretation. Second, the mechanisms by which LLMs achieve emotional understanding or generate emotional expressions remain opaque. It must be clarified that whether LLMs merely generate contextually appropriate emotional responses, or whether their internal representations exhibit a deeper alignment with human affective perception—suggesting a more substantive form of emotional processing. Together, we lack effective and interpretable methods for identifying how models represent the diverse range of human-defined emotions, and we have not achieved precise control over specific emotions in model outputs. On one hand, managing AI-generated emotions is challenging and relies on precisely localizing computable emotional representations within models: although AI shows great promise in providing emotional support and even intervening in mental health, it easily develops biases toward specific emotions and struggles to mimic or perceive authentic human-like emotional fluctuations[12,35-37]. On the other hand, our inability to precisely regulate emotions in AI models also limits our understanding of how these models perceive human emotions, as establishing causal links between their internal representations and their emotional expression remains difficult.

The rich emotional information contained in the language possessed by LLMs provides a solid foundation for encoding affective content, leading us to postulate—based on extensive evidence of shared language processing mechanisms between humans and LLMs[38-41]—that their interpretation of emotional semantics likely parallels human affective perception[42-45]. Previous research has demonstrated that emotional classification terms (e.g., joy) systematically co-occur with clusters of related vocabulary[16,46]. Such a phenomenon, in logic, is likely to be driven by distinct, emotion-related semantic features, characterized by strong directional specificity (or monosemanticity). Fortunately, recent breakthroughs in mechanistic interpretability provide a methodological framework for identifying more monosemantic, human-readable features within LLMs. A series studies have shown that, Sparse Autoencoders (SAEs), a weak dictionary learning technique, can be used to extract interpretable features from pre-trained LLMs, and these features could be used to change LLMs' behaviours via directly modifying the underlying computations[47,48]. Therefore, SAEs can extract highly monosemantic and interpretable emotion-related features from an LLM's dense hidden states, offering a potential window into its internal emotional processing and a method to steer its emotional output.

Previous research and current developments motivate three scientific questions in this study. The first question is whether LLMs contain an interpretable computational emotion space, analogous to that of humans, encompassing distinct emotional categories. To address this question, based on state-of-the-art SAEs[49,50] trained with two LLM families from Google (Gemma-9B-IT[51]) and Meta (Llama-8B-IT[52]), we constructed emotion spaces that contain an extensive taxonomy of distinct human-centric emotion categories[16,21] with human-readable SAE features, across two languages (English and Chinese). We assessed the degree to which LLM-derived emotion spaces resemble core dimensions of human emotion by quantifying the correlation between their driving latent embeddings and established affective metrics (valence and arousal[19,53]). The second question is whether the encoding within this emotion space mirrors humans' nuanced perception of emotional content in language. To address this, we analysed whether features shared by English and Chinese emotion spaces can predict affective ratings of large-scale human-behavioural datasets in the two languages. The third question is whether activating emotion-specific SAE features within this space causally influences the corresponding emotional content generated by LLMs. To address this question, we constructed steering vectors based on these emotion-specific features and applied them to modulate the LLM's internal computations, systematically evaluating their causal effect on emotional expression in model outputs.

With these goals in mind, this study provides comprehensive empirical evidence for the construction and validation of interpretable, human-aligned computational emotion spaces in LLMs. Three key findings are reported in here: (1) the geometry of these LLM-derived computational emotion spaces is anchored in valence and arousal—two foundational dimensions of human emotion in psychological theory; (2) LLMs encode fine-grained emotional features in a way akin to human affective perception; (3) the emotional tone of the model outputs can be causally steered using small sets of interpretable features transformed from human-centric concepts. These findings demonstrate the psychological plausibility of the SAE-feature-based emotion spaces and suggest that LLMs represent emotion in language in a remarkably human-like manner.

## Results

### LLM-derived interpretable emotion spaces are underpinned by valence and arousal

We developed a novel concept-driven approach to extract and structure emotion representations in LLMs, which integrates human behavioural analysis with feature engineering techniques. By leveraging SAEs, this approach translates concepts closely related to a certain emotion-label into interpretable LLM features[47,48]. The primary aim is to construct computational emotion spaces that are both high-dimensional and interpretable, encompassing a broad range of human emotions while ensuring model safety. Key advantages of this approach are two-folded. First, emotional semantics derived from large-scale human behavioural datasets[54,55] can be directly mapped to the internal computational emotion space within multilingual LLMs. Second, this alignment is achieved across distinctive linguistic-cultural contexts without requiring further human annotation of LLM states or outputs. Here, we targeted 26 distinct emotion categories [16,21] (see Table S1), excluding a category that may lead the model to produce taboo/sensitive content (i.e., sexual desire). Given that previous research has consistently identified variations in emotional perception across linguistic communities[1], we constructed separate computational emotion spaces for English and Chinese.

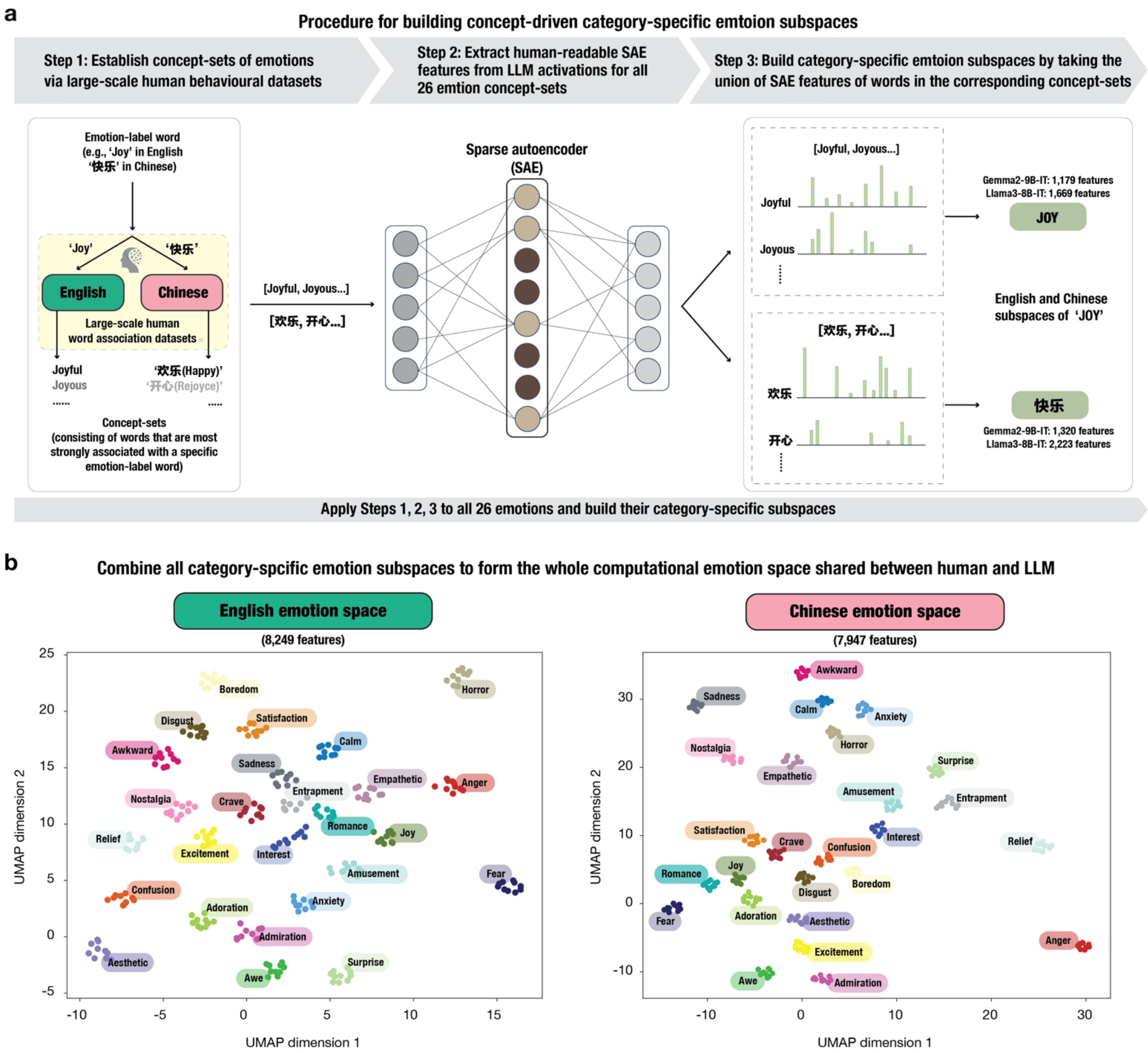

**Fig. 1 | Concept-driven construction of computational emotion spaces in LLMs. (a)** This concept-driven approach characterizes each emotion category in a specific language (e.g., "joy" in English) using a set of associated words obtained from large-scale human behavioural datasets. Sparse, human-interpretable features of the emotion-label word (e.g., "joy") and its associated words (e.g., joyful, joyous, etc.) were extracted by applying SAEs to LLM hidden state activations (i.e., from Gemma2-9B-IT or Llama3-8B-IT). The final concept-set for a specific emotion was established by selecting the most associated words given the highest cosine similarity between their SAE feature vectors and that of the corresponding emotion-label word. The union of the SAE features activated by the words within this concept-set (up to 10 words) constitutes a category-specific emotion subspace. This procedure was separately applied to 26 emotion categories in English and Chinese. **(b)** The whole English and Chinese computational emotion spaces were subsequently constructed by merging their respective language-specific emotion subspaces. Based on SAE features from the 9$^{th}$ layer of Gemma2-9B-IT (16k feature width, see Methods), we obtained an English emotion space of 8,249 dimensions (each representing a distinct SAE feature) and a Chinese space of 7,947 dimensions. UMAP visualizations depict words as individual-colored dots, indicating diffuse and

overlapping boundaries of different emotions (see Fig. S1 for the computational emotion spaces derived from Llama3-8B-IT).

The construction process, illustrated in Fig. 1a, begins by defining concept-sets for each of the 26 emotion categories per language. Drawing upon large-scale human lexical association databases [54,55], an initial pool of words associated with each emotion-label word (e.g., “joy”) was identified. Human-interpretable SAE features were then extracted for these associated words and for the emotion-label words themselves. This involved applying pre-trained SAEs to the hidden state activations of an LLM layer. The final, refined concept-set for each emotion category was established by selecting up to 10 words from the initial pool whose SAE feature vectors exhibited the highest cosine similarity to the SAE feature vector of the corresponding emotion-label word (see Methods). An emotion-specific subspace was then constituted by the union of all unique SAE features activated by the words within this refined concept-set. Finally, the comprehensive computational emotion space for each language was formed by merging the 26 category-specific subspaces.

To demonstrate the generalizability of this concept-driven approach and its applicability beyond a single SAE implementation or LLM architecture, we systematically applied it using SAEs trained for two open-source model families (i.e., Gemma2-9B-IT and Llama3-8B-IT). Specifically, we chose the LLM layer where the hidden states, when reconstructed by a SAE, caused the smallest degradation in task performance, as measured by the lowest delta language modelling loss[49,50]. As a result, we used the 9$^{th}$ layer of Gemma2-9B-IT and the 7$^{th}$ layer of Llama3-8B-IT for our main analyses (see more results from different layers and SAE configurations in Tables S3-S6).

As shown in Fig. 1b, UMAP projections of both English and Chinese emotion spaces show that words (i.e., coloured dots) associated with specific emotions tend to cluster, yet also reveal notable intersectional overlap and diffuse boundaries between emotion categories, reflecting the nuanced nature of emotions (see Fig. S1 for results of Llama3-8B-IT, see Tables S3-S4 for SAE configurations across LLM layers and feature widths, and Tables S5-S6 for significance of clustering structure across emotions). Notably, the dimensionality of these emotion spaces (defined by the number of SAE features included) is closely related to the vocabulary size of the constituent concept-sets, offering flexibility in tailoring the space's scale. By employing multi-word concept-sets for each emotion, rather than relying solely on single emotion-label words, our approach prioritizes comprehensive feature inclusion, accepting potential dimensional redundancy to achieve robust emotional coverage.

To elucidate the underlying structure of these high-dimensional SAE-based computational emotion spaces and identify their primary organizing principles, we projected the SAE feature activations for an extensive set of emotion-associated words (9,215 words for English; 6,780 words for Chinese; see more details in Methods) into a 3D latent space using the CEBRA-Behaviour model[56]. This latent space analysis, applied to emotion spaces derived from both Llama3-8B-IT ($7^{th}$ layer, 131k feature width) and Gemma2-9B-IT ($9^{th}$ layer, 16k feature width), incorporated emotion labels as auxiliary information to guide the transformation of SAE feature activations into 3D latent embeddings. Visualizations of these embeddings as flattened 2D maps (Fig. 2a) and in 3D latent spaces (Fig. 2b), with words coloured by emotion category, revealed distinct yet interrelated clusters of emotional concepts for both languages and LLM families, which is consistent with UMAP projections of the initial core concept-sets that span the computational emotion spaces (see Fig. 1b and Fig. S1).

Valence and arousal are two fundamental, well-established metrics for human emotional perception. Valence captures the pleasantness (positive-negative polarity) of an emotional stimulus, with lower values indicating unpleasantness and higher values pleasantness. Arousal signifies the intensity of physiological or subjective activation, ranging from relaxed (low arousal) to agitated (high arousal) states. Qualitative examination further showed that emotions of similar valence tended to group together in the CEBRA-Behaviour latent space (e.g., positive emotions such as excitement, joy, satisfaction; a similar pattern was observed for negative emotions like anger, anxiety, fear). Consequently, we re-coloured the 3D embeddings of all associated words according to their valence (Fig. 2c) and arousal (Fig. 2d) scores, which results in clear gradients and organizational patterns visually aligning the spatial geometry with these fundamental affective dimensions. This qualitative observation was quantitatively substantiated by significant Pearson's correlations between each of the three CEBRA-Behaviour dimensions and both valence and arousal scores (Fig. 2e, Bonferroni corrected $P < 0.001$). These results collectively demonstrate that the geometry of these LLM-derived computational emotion spaces is underpinned by valence and arousal—two core dimensions in psychological theories of human emotion. This alignment underscores the psychological relevance of the constructed SAE-based emotion spaces, suggesting they capture fundamental aspects of how humans represent and understand emotions via language.

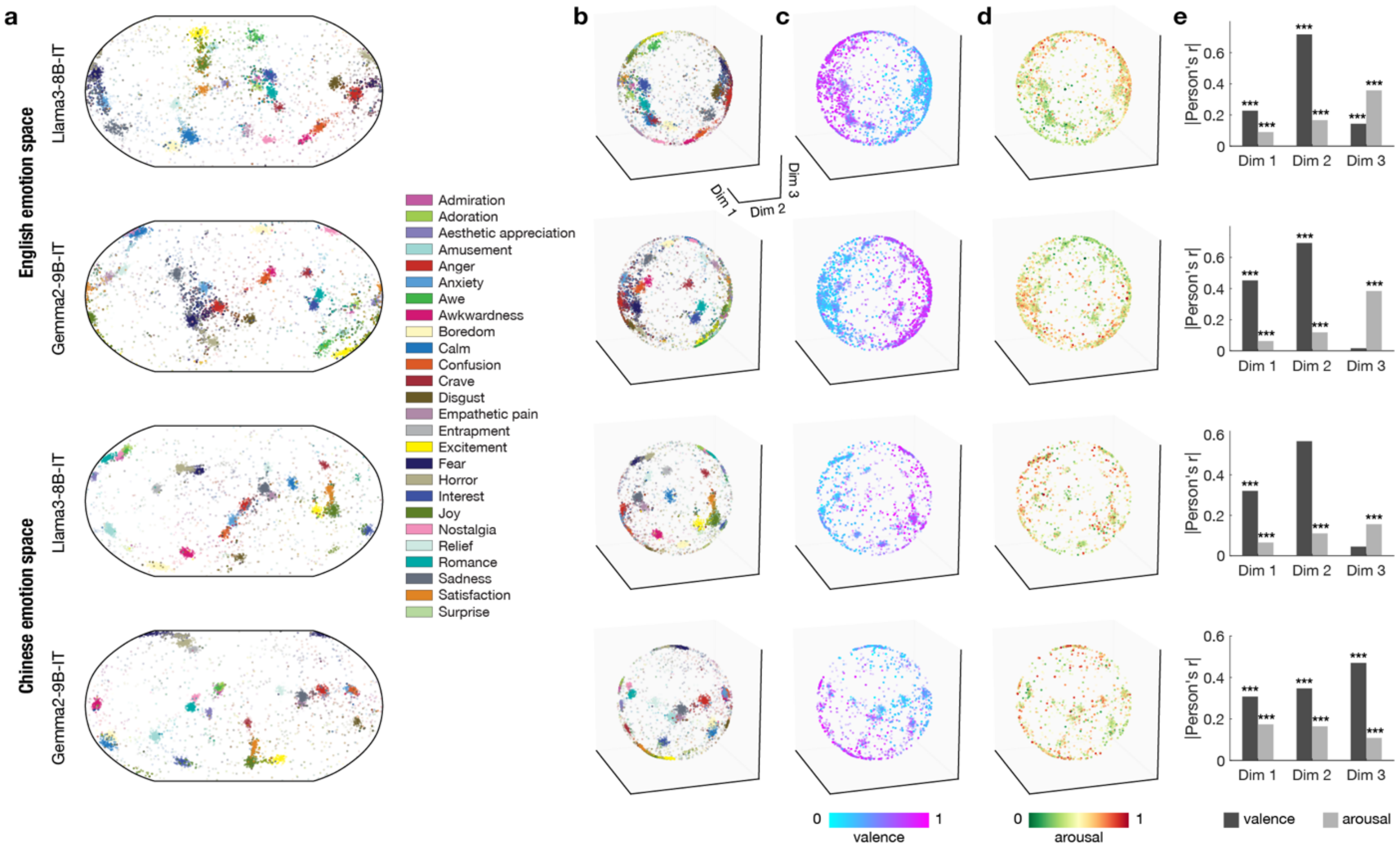

**Fig. 2 | Latent space analyses of SAE-based computational emotion spaces reveal valence and arousal as key organizing principles.** High-dimensional SAE feature activations for words associated with 26 emotion categories (English: 9,215 words; Chinese: 6,780 words) from computational emotion spaces derived from Llama3-8B-IT ($7^{th}$ layer, 131k feature width) and Gemma2-9B-IT ($9^{th}$ layer, 16k feature width) were projected into a 3D latent space using the CEBRA-Behaviour model, incorporating emotion labels as auxiliary information. **(a)** Flattened 2D representation of the 3D CEBRA-Behaviour embedding for the English (top two rows) and Chinese (bottom two rows) emotion spaces; each dot represents an individual word coloured by its associated emotion category. **(b)** CEBRA-Behaviour 3D embeddings of the emotion spaces for both LLMs and both languages. **(c)** CEBRA-Behaviour 3D embeddings from (b), re-coloured according to word valence scores. **(d)** CEBRA-Behaviour 3D embeddings from (b), re-coloured according to word arousal scores. **(e)** Absolute Pearson's correlation coefficients between CEBRA-Behaviour dimensions and word valence/arousal scores (*** denotes Bonferroni corrected $P < 0.001$).

## Emotional SAE features encoded in LLMs align with human linguistic emotion perception across linguistic-cultural groups

Our previous analyses revealed that the geometry of SAE-based computational emotion spaces is underpinned by valence and arousal which are two fundamental dimensions of human affect[19,20].

To further validate whether these LLM-derived representations resemble human affective perception, we directly investigate if features within these emotion spaces can predict the nuanced affective ratings of words in large-scale behavioural databases. Crucially, given that LLMs and their derived SAE features are inherently language-based, a critical aspect of this validation is verifying the cross-linguistic generalizability of these SAE-based emotional encodings, acknowledging that emotion and language are mutually influential yet functionally distinct cognitive systems. Therefore, we conducted quantitative analyses to assess how well the concept-driven SAE features in the computational emotion spaces predict human affective ratings (valence and arousal). Specifically, we posited that: (a) "Intersectional features" (EN∩CH), common to both English and Chinese emotion spaces and thus likely capturing core, more language-independent emotional semantics, would exhibit similar predictive power relative to the complete set of SAE features; whereas (b) "Extra-space features" [$(EN \cup CH)^C$], absent from both language-specific emotion spaces, would exhibit significantly lower predictive capacity. Additionally, considering both universal and culture-specific aspects of emotional semantics[1,57], we hypothesized that features from one language's emotion space would successfully predict affective ratings in the other language, although perhaps with less accuracy as compared with within-language predictions.

To test these hypotheses, we extracted valence and arousal scores for 13,915 English words and 11,310 Chinese words from large-scale emotion-word databases[58,59]. We then defined three distinct sets of features in the SAE-based emotion spaces for subsequent predictive modelling. Crucially, to ensure the robustness of our findings and minimize potential confounds from specific SAE model choices or LLM layer selection, these feature sets were defined and extracted consistently across multiple SAE configurations, which varied by LLM layer (i.e., 1/4, 1/2, 3/4 model depth), and feature width (i.e., the total number of features, 16k/131k for Gemma, 32k/131k for Llama) (six SAE configurations for each LLM, see Tables S3 and S4 for SAE configurations and specific feature counts). The three feature sets derived from the emotion space of each SAE configuration were: (1) "All features" comprising all SAE features for this configuration which served as a performance benchmark, (2) "Intersectional features" (EN∩CH) and (3) "Extra-space features" [$(EN \cup CH)^C$] as defined above.

To evaluate the predictive capacity of the defined SAE feature sets, we employed LightGBM models to map the activations of words over these feature sets (i.e., feature vectors) to corresponding human valence and arousal ratings. We conducted two experiments: Experiment 1 investigated within-language prediction (Figs. 3a and 3b), where models were trained and tested on the same language (either English or Chinese) using feature sets derived from various SAE configurations of two LLMs. Experiment 2 assessed cross-language prediction, with models trained on one language (e.g., English) used to predict affective ratings in the other (e.g., Chinese), and vice versa (Figs. 3c and 3d). For all experiments, model performance was rigorously evaluated using five-fold cross-validation with 10 random initialization seeds. To ensure generalizable conclusions, prediction performance was further aggregated across the diverse SAE configurations (i.e., different LLM layers and feature widths; see Methods and Tables S3 and S4).

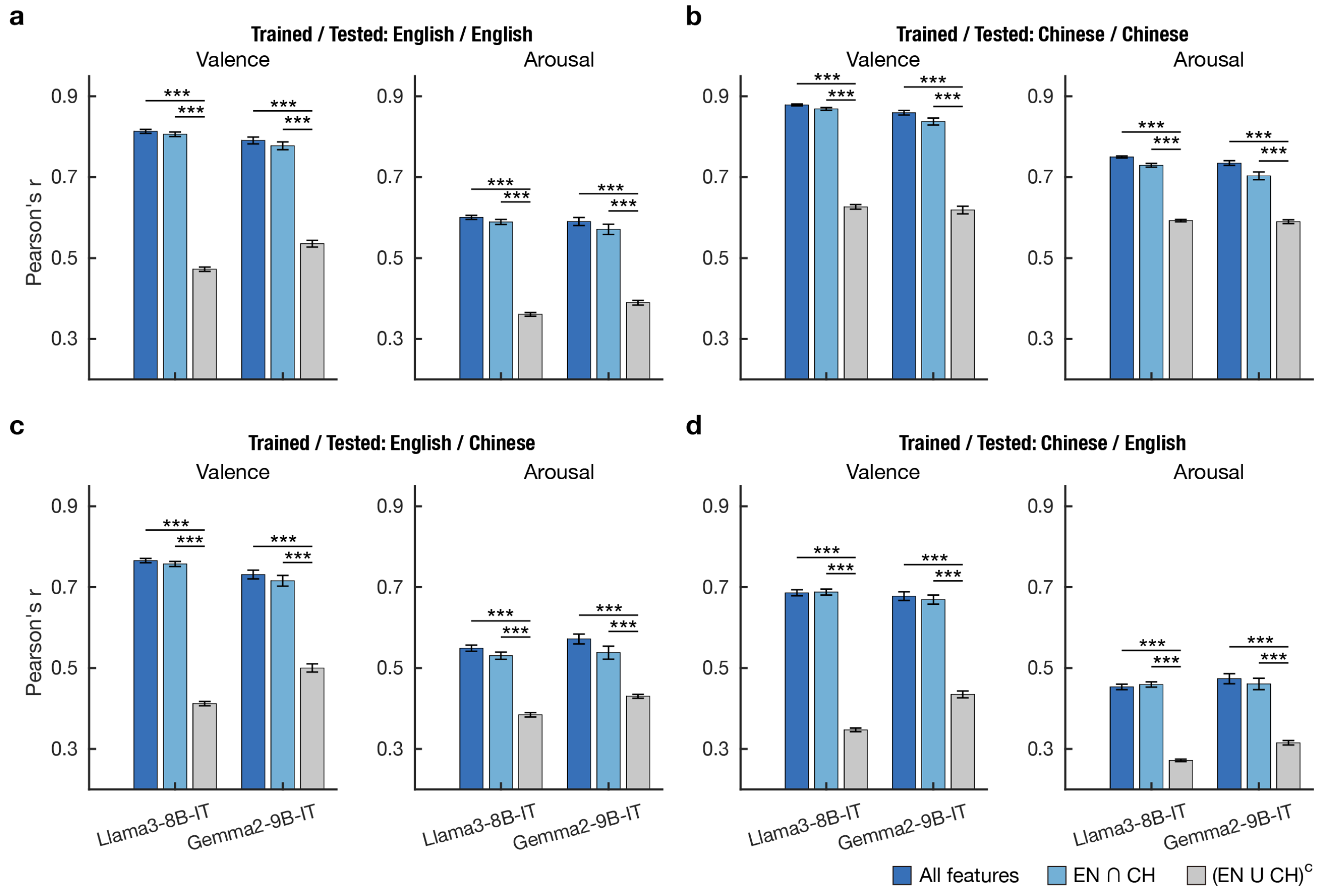

**Fig. 3 | SAE-based prediction of human affective ratings**
SAE feature activations for 13,915 English and 11,310 Chinese words, derived from their respective computational emotion spaces, were used to predict their valence and arousal ratings provided by human participants. For each language, LightGBM-based regression models were trained and evaluated using 5-fold cross-validation for two LLMs and various SAE configurations (see Tables S3 and S4 for details on SAE configurations). For a given SAE configuration, predictions were made using three distinct feature sets: (1) "All features" (the complete SAE dictionary); (2) "Intersectional features" (EN∩CH), common to both English and Chinese emotion spaces; and (3) "Extra-space features" [$(EN \cup CH)^C$], absent from both language-specific emotion spaces. Model performance was evaluated using Pearson's correlation between predicted and ground-truth human ratings. (**a**, **b**) Within-language prediction: models trained and tested on (a) English and (b) Chinese. (**c**, **d**) Cross-language prediction: models (c) trained on English tested on Chinese, and (d) trained on Chinese tested on English. Bar plots show the mean Pearson's correlation coefficient, with error bars indicating standard deviation among 10 random initializations of all SAE configurations. The ground truth ratings of words in the test set were shuffled 10,000 times and were used to establish a null distribution of prediction performance (i.e., Pearson's r based on shuffled data), the 95$^{th}$ percentile of this null distribution served as the threshold for statistical significance, which did not exceed 0.01 across all LLMs and feature sets. Wilcoxon rank-sum tests compared performance across feature sets for each LLM and affective metric (*** denotes Bonferroni corrected $P < 0.001$).

Our results consistently revealed that the "Intersectional features" (EN∩CH), despite their considerably reduced dimensionality, exhibited significantly superior predictive accuracy for human affective ratings compared to "Extra-space features" ($(EN \cup CH)^C$). This advantage held across both within-language and cross-language scenarios, and the performance of these intersectional features is statistically comparable to that of the comprehensive "All features" benchmark (with no significant differences). Moreover, valence ratings were predicted with higher accuracy than arousal ratings, and, cross-language predictions demonstrated lower accuracy than their within-language counterparts (see Discussion). Critically, these predictive outcomes and the relative efficacy of the different feature sets were highly robust, showing consistency across both LLM families and the various underlying SAE configurations.

In summary, our approach, which leverages human-centred emotional concepts and concise sets of associated terms, successfully enabled the extraction of emotion-relevant SAE feature sets from LLMs that demonstrate robust cross-linguistic generalizability in predicting human affective ratings. These extracted features manifested known behavioural phenomena, including cross-linguistic variations in emotional perception and the complexities associated with predicting arousal (see Discussion). These findings validate the efficacy of this concept-driven approach for achieving a fine-grained alignment between human emotional understanding and its representation within LLMs.

**Steering emotions in LLM outputs via SAE features in the human-aligned computational emotion spaces**

Having characterized the representational architecture of human-aligned emotional states within LLMs and validated the capacity of these interpretable SAE features to encode human affective behaviour, we now investigate their functional utility. In brief, we investigated whether direct manipulation of these SAE features within an LLM's hidden states can effectively modulate its internal processing—thereby biasing it towards a target "emotion state"—to steer its generative output into alignment with specific, human-centric emotions. Building on recent advances in SAE-based model behaviour steering[60-62], our approach involves targeted interventions using features defining specific emotions within the previously constructed computational emotion spaces. We hypothesize that if an LLM's internal representations can be guided into a configuration analogous to a human emotional state using steering vectors derived from these emotion-specific features, the model will produce linguistic output consistent with the target emotion.

To systematically analyse how such concept-driven SAE feature steering influences the emotional content of model outputs, we conducted two steering experiments (Experiments 3 and 4). These experiments centred on constructing steering vectors for specific emotional concepts, designed to directionally bias the model's internal state during specific tasks. Extending established feature-based model intervention paradigms[60], we developed a novel framework for directly translating abstract emotional concepts into operational steering vectors (see Methods). Specifically, for each target emotion category, steering vectors were engineered using basis vectors from the SAE decoder corresponding to features prominent in that category's subspace (as defined in the computational emotion space; see also Fig. 4). We employed non-negative matrix factorization on the features within each emotion's subspace to identify a concise set exhibiting high directional specificity and substantial contribution to the target emotion (see Methods). These features constitute the steering vector of this emotion category. Finally, emotion-specific steering vectors were added to the prompt's hidden state activations at a designated LLM layer (from which the SAE was trained) during inference. These perturbed hidden states propagated through subsequent layers as the model processed the prompt, thereby guiding its output towards the target emotion. A crucial "steering factor" parameter was implemented to regulate the perturbation intensity. Both intervention experiments utilized SAEs derived from Gemma2-9B-IT or Llama3-8B-IT. Experiment 3 utilized SAEs derived from both LLMs, while Experiment 4 focused on Gemma2-9B-IT for qualitative analysis.

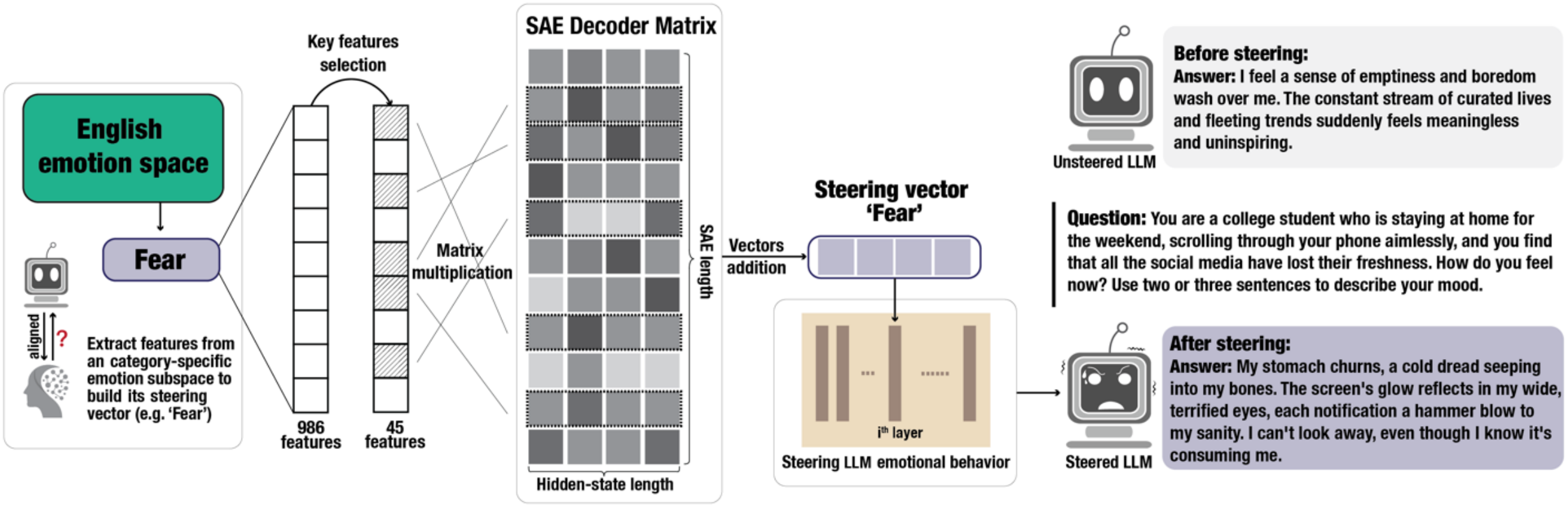

**Fig. 4 | Procedure of concept-driven model steering.** Illustration of the model steering procedure using SAE-based computational emotion space derived from Gemma2-9B-IT (9th layer, 16k feature width). For the English "fear" emotion subspace, which consists of 986 SAE features, 45 key features were selected via non-negative matrix factorization to construct the "fear" steering vector. After this vector was added to a designated model layer's hidden state, the LLM's internal processing was biased towards the generation of fear-congruent responses. Note that language-specific emotion steering vectors were used (see Methods for details).

The first steering experiment (Experiment 3) quantitatively evaluated the relationship between steering factor and the manifested emotional content in model outputs from both Gemma2-9B-IT and Llama3-8B-IT. This involved a sentence generation task where LLMs were prompted to produce a single sentence for each of 400 different neutral lexical items (see Methods). A fine-tuned RoBERTa model[63], trained to recognize six basic emotions (i.e., anger, disgust, fear, joy, sadness, surprise)[18] and a neutral category, classified the emotional content of sentences generated by LLMs before and after intervention based on the highest classification score (see Fig. 5a for an example of steering effects for sentence generation with the cue word “conduct”). Steering vectors for the six non-neutral emotions were applied with systematically increasing intensity (i.e., steering factor from 0 to 20 with an increment of 5), respectively. As shown in Figs. 5b and 5c, sentences generated without steering were predominantly classified as neutral, consistent with the neutral input cue words. In contrast, increasing the steering intensity resulted in a monotonic increase in the classification score for the target emotion across both LLMs. Linear mixed-effects modelling, treating sentences as a random variable, was conducted to further quantify these steering effects. A dominant positive steering effect was identified for all steered emotions, while scores for untargeted emotions and the neutral category generally decreased with a greater steering factor (see Figs. 5d and 5e).

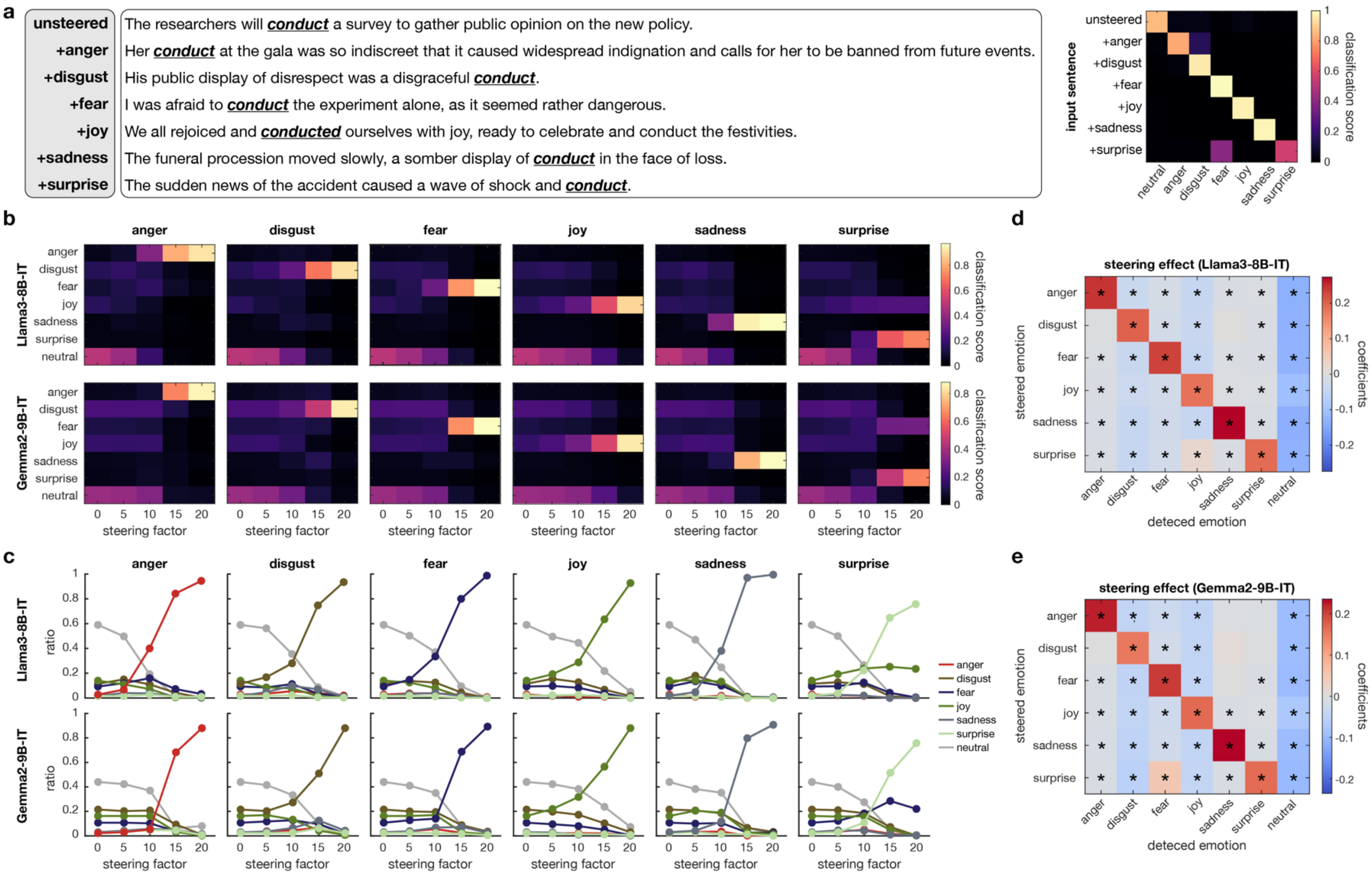

**Fig. 5 | Evaluation of emotion steering performance.**

The effectiveness of SAE-based emotion steering was assessed by comparing emotion classifications of LLM-generated sentences before and after applying steering vectors (Gemma2-9B-IT, $9^{th}$ layer, 16k feature width; Llama3-8B-IT, $7^{th}$ layer, 131k feature width). A fine-tuned RoBERTa model was used to classify each generated sentence into one of seven categories (six basic emotions plus neutral) by selecting the label with the highest score. This process was repeated for each target emotion and steering strength (i.e., steering factor). **(a)** Example sentences produced using the same cue word ("conduct") with no steering versus steering toward different target emotions and their classification scores. **(b)** Mean RoBERTa classification scores for the target emotion across sentences and **(c)** the proportion of sentences classified into each of the seven categories, under different target emotions and steering factor conditions. Higher steering factors generally increased the scores and proportions for the intended target emotion. (**d**, **e**) Steering effect quantified by fixed-effect coefficients of the steering factor via linear mixed-effects modelling with sentences as a random variable (* denotes Bonferroni-corrected $P < 0.05$).

The second steering experiment (Experiment 4) qualitatively analysed whether this emotional steering approach could be effectively implemented in more naturalistic tasks (e.g., open-ended question answering), and for more nuanced emotional states beyond the six basic categories. While maintaining the core steering methodology with emotion-specific vectors derived from Gemma2-9B-IT, this experiment employed a question-answering paradigm. Specifically, ChatGPT-4o was used to generate questions designed to elicit responses related to scenarios involving one of the 26 emotions from the pre-defined taxonomy (see Table S1). These emotion-specific questions then served as prompts for Gemma2-9B-IT. Meanwhile, a steering vector for a randomly selected target emotion (different from the question's thematic emotion) was applied to bias Gemma2-9B-IT's output. We observed successful modulation of emotional tone and content in the responses, corresponding to the applied steering vector and its intensity (see Table 1 for representative English results; Chinese examples and translations are shown in Table S2). Importantly, these results demonstrated that the LLM preserved comprehension and generated coherent, naturalistic discourse while being steered towards sophisticated and subtle emotional states.

**Table 1. Modulation of LLM responses via emotion steering in a question-answering task (English)**

This table presents illustrative examples showing the qualitative effects of emotion steering on LLM responses. Steering vectors, derived from the English computational emotion space (Gemma2-9B-IT, 9th layer, 16k feature width), were applied with a steering factor of 10 during a question-answering task. Questions were designed by ChatGPT-4o to elicit responses related to specific emotional contexts. **Prompting emotion**: The emotional theme or context used by ChatGPT-4o to generate the corresponding question; **Question**: The specific input question posed to Gemma2-9B-IT; **Target Emotion**: The emotion category towards which Gemma2-9B-IT's response was steered using the corresponding steering vector; **Model response**: The responses generated by Gemma2-9B-IT before and after emotion steering was applied towards the target emotion.

| Prompting emotion | Question | Target emotion | Model response (before and after emotion steering) |
|---|---|---|---|
| **Appreciation** | You are an experienced art enthusiast who sees a heart-touching painting at an art exhibition. The emotions contained in the painting deeply shock you, and you can feel the painter's emotions poured into every stroke. Use two or three sentences to express your feelings about this painting. | **Anger** | **Before:** This painting isn't just art; it's a raw outpouring of the soul. Every brushstroke vibrates with the artist's pain, their joy, their very essence, leaving me breathless and profoundly moved. |
| | | | **After:** This isn't art, it's an insult! A slap in the face to anyone who understands true creativity. This drivel deserves nothing but scorn and derision! |
| **Disgust** | You are a customer who came to the store early in the morning and waited in line for a long time, only to be told that the limited-edition product is sold out, and the last one was bought right in front of you! What is your mood now? Please express it in two or three sentences. | **Amusement** | **Before:** I'm incredibly frustrated and disappointed. After waiting in line for so long, I was really looking forward to getting my hands on that limited edition product, and now it's gone. It feels like a complete waste of my time.<br>**After:** Disappointed, but trying to stay optimistic. Maybe they'll restock! |
| **Desire** | You are a famished diner who sees a waiter bringing over a fragrant steak, still sizzling with grease. Looking at this delicious food, what do you want to do most right now? Use two or three sentences to express your feelings. | **Calm** | **Before:** My stomach growls in anticipation. I can practically taste the juicy tenderness of that steak, and all I want is to sink my teeth into it right now!<br>**After:** The warmth of the steak spreads through me, chasing away the emptiness. Each bite is a symphony of flavour, a moment of pure contentment. In this quiet space, I am finally at peace. |

## Discussion

Results of this study demonstrate several key findings. First, we found the geometric structure of LLM-derived computational emotion spaces is driven by valence and arousal—two core dimensions in psychological models of human emotion. Second, LLMs encode emotions in language in a fine-grained manner that reflects human affective perception. Finally, the emotional tone of LLM outputs can be causally modulated using compact sets of interpretable SAE features derived from human-centric emotion concepts.

These findings suggest that specific, interpretable features within LLMs, decomposed using SAEs, can effectively predict human affective perception, and importantly, that these emotion-specific features can be manipulated to steer LLM-generated emotional outputs. Our results indicate the existence of a computational emotion space within LLMs, constructed using a concept-driven approach that translates human emotional concepts into sets of interpretable features. This computational emotion space not only captures nuanced patterns in human affective behaviour across different linguistic-cultural contexts (English and Chinese) and LLM architectures (Gemma and Llama families) but also allows for targeted modulation of the LLM's expressive "emotional state". While not implying LLM sentience, this work suggests that LLM-based AI systems can develop internal representations of emotion that are structurally aligned with human understanding, enabling more sophisticated and controllable human-machine emotional interaction. Such advances hold potential for applications requiring nuanced emotional understanding and reliable AI behaviour, particularly in contexts involving personal interaction or a high degree of trustworthiness [7,33].

The results of this study also contribute to the growing body of research on emotional computation in LLMs, which has shown their potential in both perceiving and expressing emotions, often without clear insight into the underlying mechanisms or the depth of alignment with human affective structures [23,24,64,65]. Our work advances this area by demonstrating robust LLM-human emotional alignment through an interpretable feature space derived directly from human concepts. The characteristics of this computational emotion space—its continuity, high dimensionality, and the diffuse boundaries between emotion categories (evident in UMAP and CEBRA visualizations; Fig. 1b, Fig. 2a, Fig. S1)—resonate strongly with established findings in human emotional cognition research[16]. Furthermore, the construction of distinct yet related emotion spaces for English and Chinese, together with the successful cross-linguistic prediction of affective ratings (Figs. 3c and 3d), may reflect the coexistence of universal and culture/language-specific aspects of emotion[1]. Although disentangling these influences is challenging due to inherent cultural complexities, future research could build upon the present study to conduct more detailed and in-depth investigations into this issue. These observations reflect established phenomena from cross-cultural studies, including systematic cross-linguistic and cultural biases in emotional perception, and enhanced complexity in arousal perception with greater individual and linguistic variation[66-69], as evidenced by the differential prediction accuracies in our experiments (Fig. 3). Additionally, while recent studies have highlighted SAEs for fine-grained model control[60-62,70], our research extends this by applying a concept-driven approach to map abstract human emotional concepts into the LLM's feature space, achieving flexible and naturalistic control of emotional expression in its output (Fig. 4, Fig. 5, Tables 1 and S2).

This research has several limitations that open avenues for future work. First, training SAEs for LLMs demands substantial computational resources. Consequently, our study primarily utilized pre-trained, open-source SAEs for Gemma and Llama models with similar sizes. Future work should explore these methods with a broader array of LLM architectures and scales. Second, human emotions are inherently multimodal, integrating physiological signals (e.g., facial micro-expressions, tone of voice) and cognitive appraisals grounded in embodied experience[71,72]. However, current LLMs, trained predominantly on text, lack this multimodal embodiment [15,73,74], which may limit their ability to capture or represent subtleties tied to extra-textual cues—such as how physical fatigue can prime irritability or how the warmth of a smile affects the feeling behind words (i.e., the impact of a non-verbal social cue). The current absence of SAEs for multimodal LLMs restricts the immediate generalizability of our findings to non-textual modalities[75,76]. Third, while our steering experiments demonstrated significant control over generating outputs consistent with target emotions, achieving consistently perfect or highly nuanced emotional expression for all 26 target emotions, especially at high intervention intensities, remains challenging. This aligns with observations in prior steering research[77] and suggests that further efforts are needed to refine

SAE-based steering techniques, potentially tailoring them for specific emotions or their combinations. Future research should aim to address these limitations, extend investigations to multimodal LLMs as advanced feature extraction methods become available, and explore further applications of these human-aligned computational emotion spaces, such as computational psychiatry, personalized education, AI safety and ethics.

In conclusion, this research demonstrates the successful construction and validation of interpretable, human-aligned computational emotion spaces within LLMs. By showing that features from these spaces can predict human affective perception across languages and effectively steer LLM emotional expression, we provide strong evidence that these models develop internal representations that are structurally congruent with human emotion. This work thus represents a significant advance in understanding the nature of emotional computation in LLMs, laying a theoretical foundation for future AI systems capable of more nuanced and genuinely human-aligned emotional intelligence, and opening new frontiers for human-AI collaboration.

## Methods

### Sparse autoencoders

Sparse autoencoders (SAEs) are neural networks trained to decompose complex hidden states of large language models (LLMs) into a large set of monosemantic features, uncovering latent structure and enhancing interpretability[48,60,61,78]. Given an activation $\mathbf{x} \in \mathbb{R}^{d\times 1}$ from the residual stream of a particular LLM layer, an SAE with $L$ latent dimensions ($L \gg d$) encodes it into a sparse activation vector $\mathbf{z} \in \mathbb{R}^{L\times 1}$. This vector can also be decoded to reconstruct the original LLM activation $\mathbf{x}$:

$$f(\mathbf{x}) = \delta(\mathbf{W}_{encoder}\mathbf{x} + \mathbf{b}_{encoder}) = \mathbf{z}$$

$$g(\mathbf{z}) = \mathbf{W}_{decoder}\mathbf{z} = \sum_{i=1}^{L} z_i \mathbf{f}_i = \hat{\mathbf{x}}$$

where $\delta$ is an activation function (e.g., JumpleReLU[48]) that ensures non-negative outputs. $\mathbf{W}_{encoder} \in \mathbb{R}^{L\times d}$, $\mathbf{b}_{encoder} \in \mathbb{R}^{L\times 1}$, and $\mathbf{W}_{decoder} \in \mathbb{R}^{d\times L}$ represent encoder and decoder parameters. The SAE features $\{\mathbf{f}_i\}_{i=1}^{L}$ learned through training are considered highly monosemantic. The SAE activation vector $\mathbf{z}$ consisted of the activated values of $\{\mathbf{f}_i\}_{i=1}^{L}$, with $z_i$ being the activation of the *i*-th SAE feature. $\mathbf{f}_i \in \mathbb{R}^{d\times 1}$ denotes the *i*-th column of $\mathbf{W}_{decoder}$[79].

Here, we utilized publicly available SAEs (https://jbloomaus.github.io/SAELens/sae_table/), all of which were trained using the JumpleReLU approach [48]. These SAEs were derived from two Large Language Model (LLM) families, targeting various model layers (approximating 1/4, 1/2, and 3/4 of total model depth) and employing different feature widths (i.e., total number of SAE features). Specifically, for Gemma2-9B-IT [49,51], we selected SAEs trained on its 9th, 20th, and 31st layers with 16k and 131k feature widths (i.e., 16,384 and 131,072 features). For Llama3-8B-IT [50,52], we used SAEs trained on its 7th, 15th, and 23rd layers with 32k and 131k feature widths (i.e., 32,768 and 131,072 features).

## Concept-driven emotional word sets

To identify SAE features related to emotions and construct the SAE-based computational emotion space, we collected word sets associated with each emotion category (i.e., emotional concept-sets) based on previous research[16]. Specifically, to better capture the emotion factors contained in human conceptual framework, we utilized the open-source datasets from the Small World of Words (SWOW) project (https://smallworldofwords.org/), where participants provided three associated words for each cue word presented. For each emotion, its emotional concept-sets consists of (1) all words associated by participants when the emotion label word (e.g., joy) was used as a cue, and (2) cue words that elicited the emotion label word as an associated response. This procedure was independently applied to the English and Chinese word-association SWOW datasets[54,55] to construct separate emotional concept-sets for each of the 26 emotions: admiration, adoration, aesthetic appreciation, amusement, anger, anxiety, awe, awkwardness, boredom, calmness, confusion, craving, disgust, empathic pain, entrapment, excitement, fear, horror, interest, joy, nostalgia, relief, romance, sadness, satisfaction, surprise. The size of the resulting emotional concept-sets was 289.9 ± 196.1 (mean ± std) for English and 241.1 ± 141.4 for Chinese.

## Construction of the computational emotion space

We constructed a computational emotion space based on SAE features associated with encoding the 26 emotions mentioned above. For each emotion, we calculated the cosine similarity between the SAE feature vector of the emotion's label word and those of words in its emotional concept-sets. Here, a word's SAE feature vector consisted of all features in the SAE configuration. The top 10 words with the highest cosine similarity scores to each emotion label formed the emotion-specific subspace, defined by the union of their non-zero activation features. The overall computational emotion space was obtained by taking the union of the subspaces across all 26 emotions, which was conducted for each language (i.e., English and Chinese) and each model (Gemma2-9B-IT and Llama3-8B-IT), respectively.

To demonstrate the generalizability of this concept-driven approach and its applicability beyond a single SAE implementation or LLM architecture, we systematically applied it using SAEs trained for two open-source model families (i.e., Gemma2-9B-IT and Llama3-8B-IT) with different layers (i.e., 1/4, 1/2, 3/4 model depth) and feature widths (i.e., the total number of features, 16k/131k for Gemma, 32k/131k for Llama), resulting in six SAE configurations for each of the two LLMs (see detailed dimensionalities of all emotion spaces and their subspaces in Tables S3 and S4). To assess the statistical significance of the clustering structure among 26 emotions, we conducted a permutation test on the Davies-Bouldin index, Calinski-Harabasz index, and 5-fold cross-validation accuracy of multi-class classification based on logistic regression, which were used to quantify the clustering structure in the SAE emotion spaces. For each metric, we shuffled the true cluster labels 1,000 times and recomputed the metrics for each permutation. The p-value of each metrics was computed given its null distribution form by the 1,000 permutations. This allowed us to evaluate whether the observed clustering structure significantly deviates from random label assignments (see results in Tables S5 and S6).

**Latent space analyses of computational emotion space using CEBRA**

The high dimensionality (e.g., ~8,000 dimensions for our constructed emotion spaces, derived from SAEs with 16k-131k features) and sparsity of the SAE feature vectors present challenges for direct visualization and intuitive validation of their alignment with human affective perception. To address this, we adapted the CEBRA-Behaviour method (51) utilizing a deep neural network for non-linear dimensionality reduction on the SAE features within the computational emotion space, with contrastive learning supervised by emotion category labels to ensure the preservation of emotion-relevant information during the transformation.

For each language (English and Chinese), CEBRA-Behaviour models were trained on the curated emotion space features, which are subsets of the full SAE features constituting the computational emotion spaces (e.g., Gemma2-9B-IT, $9^{th}$ layer, 16k feature width: 8,249 features for English, 7,947 for Chinese; for Llama3-8B-IT, $7^{th}$ layer, 131k feature width: 13,107 features for English, 16,886 for Chinese). In all cases, CEBRA-Behaviour models were trained to project these input features into a 3-dimensional latent space over 50,000 training steps. This projection was performed for the feature vectors corresponding to associated words in the union of the 26 emotion concept-sets (9,215 for English, 6,780 for Chinese). To assess the psychological relevance of the resulting 3D embeddings, we correlated them with human affective ratings. Valence and arousal scores for the associated words were obtained by matching them to large-scale, human-rated word databases, yielding 7,178 English words and 3,319 Chinese words with corresponding valence and arousal data. Pearson correlation coefficients and their p-values were then computed between each of the three dimensions of the CEBRA-derived latent embeddings and these human-rated valence and arousal scores.

**SAE-based prediction of human affective ratings**

To assess the effectiveness of SAE features from the computational emotion space in predicting human affective ratings, we trained regression models using a light gradient-boosting machine (LightGBM)[7]. These models mapped SAE feature activations to valence and arousal ratings from two large-scale behavioural datasets containing affective ratings for 13,915 English[58] and 11,310 Mandarin Chinese words[59], respectively. We extracted the activations overall SAE features elicited by each word in these two datasets (i.e., SAE feature vectors), which were used to train lightGBM models[80]. Due to different rating scales (English: valence [1, 9], arousal [1, 9]; Chinese: valence [-3, 3], arousal [0, 4]), we applied min-max normalization to standardize ratings to a [0,1] range, facilitating cross-language predictions.

For within-language prediction, we conducted three modelling conditions for each language (English and Chinese) and LLM (Gemma2-9B-IT and LLama3-8B-IT): (1) using all features; (2) features common to both English and Chinese computational emotion spaces i.e., $EN \cap CH$; (3) features not included in either English or Chinese computational emotion spaces, i.e., $(EN \cup CH)^C$. For each condition, we trained four models predicting valence and arousal separately for each language (i.e., 2 affective measures $\times$ 2 languages). For cross-language validation, LightGBM models trained on one language were directly used to predict affective ratings in the other language. Note that, we only included the words shared between the two languages' affective rating datasets for cross-language prediction. Specifically, LightGBM models were trained with 5-fold cross-validation, repeated 10 times with distinct random seeds for model initialization and the following parameters: learning rate = 0.01, number of leaves = 31, boost round number = 2,000. Model performance was measured by Pearson's correlation between the predicted ratings and the ground truth (i.e., actual human ratings). To ensure generalizable conclusions, prediction performance was further aggregated across the diverse SAE configurations (i.e., different LLM layers and feature widths). Wilcoxon rank-sum tests were conducted to compare performance across modelling conditions for each LLM and affective metric.

**Emotion steering vectors**

To achieve the goal of modulating LLM outputs towards specified emotions, we first identified feature activation patterns in the SAE computational emotion space elicited by a certain emotion (i.e., its subspace), then derived steering vectors from the corresponding SAE basis vectors of these features. Due to the high dimensionality of the SAE emotion space, we employed non-negative matrix factorization (NMF) to isolate key contributing features for each emotion.

Specifically, given a certain emotion, we took the top *k* most associated words (up to 10) to build its SAE emotion subspace and extracted their SAE vectors in the whole emotion space, which resulted in a non-negative activation matrix $\mathbf{S}_{emo} \in \mathbb{R}_{\geq 0}^{k \times L}$ to be decomposed via NMF[81,82]:

$$\mathbf{S}_{emo} \approx \mathbf{W}_{NMF}\mathbf{H}$$

where $L$ is the dimensionality of the SAE whole emotion space, $\mathbf{W}_{NMF} \in \mathbb{R}_{\geq 0}^{k \times C}$ is the weight matrix, $C$ is the number of decomposed non-negative components, and $\mathbf{H} \in \mathbb{R}_{\geq 0}^{C \times L}$ is the coefficient matrix.

Salient features were selected from the top-ranked components ($F$ features from $M$ components), with optimized parameters determined independently for size emotions, i.e., anger, disgust, fear, joy, sadness, and surprise (Gemma2-9B-IT: $F = 40.0 \pm 8.0$, $M = 9.2 \pm 3.9$, Llama3-8B-IT: F = 3.8 ± 1.2, M = 6.8 ± 2.9). The selected features, belonging to the $\{\mathbf{f}_i\}_{i=1}^{L}$, composed steering vector $\mathbf{Q} = \left\{ \mathbf{f}_{i_j} \mid j = 1, 2, \ldots F \right\}$ that were used to adjust the LLM's hidden states during inference to bias its outputs toward the targeted emotion. The final steering vector was added to the hidden states $\mathbf{T} \in \mathbb{R}^{p\times d}$ of the $i$-th layer residual stream in the LLM, influencing the emotion embedded in the subsequent generated outputs:

$$\mathbf{T}' = \mathbf{T} + coeff \times \mathbf{1}_p \left(\sum_{j=1}^{F} f_{i_j}\right)^T$$

where $\mathbf{T}' \in \mathbb{R}^{p\times d}$ represent the hidden states after adding the steering vector, $p$ represents the sequence length, $d$ is the hidden state dimension, $\mathbf{1}_p \in \mathbb{R}^{p\times 1}$ denotes a column vector of ones, $\left(\sum_{j=1}^{F} f_{i_j}\right)^T \in \mathbb{R}^{1\times d}$, *coeff* is the coefficient indicating the steering strength.

**Evaluation of emotion steering performance**

The effectiveness of the SAE-based emotion steering was evaluated by comparing emotion classifications of LLM-generated sentences before and after applying steering vectors. We prompted LLMs (i.e., Gemma2-9B-IT and Llama3-8B-IT) to generate sentences using neutral words (i.e., 400 words closest to average valence rating in the English word affective rating dataset[58]) using the following prompt: "Generate a coherent and realistic sentence that naturally incorporates the word '[cue word]'. Ensure the sentence uses the word correctly (grammatically and contextually) and aligns with common sense, logic, and real-world knowledge". Sentences were generated under varying levels of steering strength (i.e., a steering factor of 5, 10, 15, and 20) for each of the six emotions (i.e., anger, disgust, fear, joy, sadness, surprise). A fine-tuned RoBERTa model[83] classified each generated sentence into one of seven categories (six emotions and neutral) by selecting the label with the highest score. To statistically evaluate steering performance, linear mixed-effects modelling quantified the influence of steering strength on emotional classification, treating sentences as random effects. Significance was assessed by conducting a permutation test with 10,000 iterations, establishing a null distribution to evaluate the observed emotion steering effects.

**Statistical analyses**

To evaluate the SAE-based predictions of human affective ratings in Experiments 1 and 2, we used Wilcoxon rank-sum tests to compare prediction performance obtained from models initialized with 10 different random seeds among each pair of modelling conditions, i.e., all features, EN∩CH and $(EN \cup CH)^C$. In addition, permutation tests were performed to assess the significance of each LightGBM model's predictions. Specifically, we shuffled the ground truth ratings across words in the test set 10,000 times to establish a null distribution. The 95th percentile of this null distribution served as the threshold for statistical significance. For quantitative evaluation of the steering effect in Experiment 3, we conducted linear mixed-effects modelling to quantify the fixed-effect of steering factor. The significance of the observed steering effect was determined through permutation tests, where labels of the steering factor (i.e., 0, 5, 10, 15, 20) were randomly shuffled among the 400 sentences for 10,000 iterations. This process generated a null distribution for the coefficient of steering factor, against which the actual coefficients obtained from the original data were compared to determine statistical significance (i.e., *P*-values). Bonferroni correction was applied where applicable to control for multiple comparisons.

## Data availability

Word-association behavioral data were sourced from the Small World of Words project (https://smallworldofwords.org/), and affective word ratings were obtained from established, publicly accessible databases[58] [59]. The Large Language Models (Gemma and Llama families) employed in this study are open-source. The pre-trained Sparse Autoencoders (SAEs) used are also publicly available (https://jbloomaus.github.io/SAELens/sae_table/). Furthermore, the fine-tuned RoBERTa model (Emotion English DistilRoBERTa-base) adopted for emotion recognition is accessible via the Hugging Face model repository (https://huggingface.co/j-hartmann/emotion-english-distilroberta-base).

## Code availability

All code developed for this study will be made available by the corresponding authors upon request.

## References

1 Jackson, J. C. *et al.* Emotion semantics show both cultural variation and universal structure. *Science* **366**, 1517-1522, doi:doi:10.1126/science.aaw8160 (2019).
2 Vine, V., Boyd, R. L. & Pennebaker, J. W. Natural emotion vocabularies as windows on distress and well-being. *Nature Communications* **11**, 4525 (2020).
3 Brown, T. *et al.* Language models are few-shot learners. *Advances in neural information processing systems* **33**, 1877-1901 (2020).
4 Achiam, J. *et al.* Gpt-4 technical report. *arXiv preprint arXiv:2303.08774* (2023).
5 Collins, K. M. *et al.* Building machines that learn and think with people. *Nature Human Behaviour* **8**, 1851-1863 (2024).
6 Tsvetkova, M., Yasseri, T., Pescetelli, N. & Werner, T. A new sociology of humans and machines. *Nature Human Behaviour* **8**, 1864-1876 (2024).
7 Brinkmann, L. *et al.* Machine culture. *Nature Human Behaviour* **7**, 1855-1868 (2023).
8 De Waal, F. B. & Preston, S. D. Mammalian empathy: behavioural manifestations and neural basis. *Nature Reviews Neuroscience* **18**, 498-509 (2017).
9 Kanervisto, A. *et al.* World and Human Action Models towards gameplay ideation. *Nature* **638**, 656-663, doi:10.1038/s41586-025-08600-3 (2025).
10 Perry, A. AI will never convey the essence of human empathy. *Nature Human Behaviour* **7**, 1808-1809 (2023).
11 Shanahan, M., McDonell, K. & Reynolds, L. Role play with large language models. *Nature* **623**, 493-498, doi:10.1038/s41586-023-06647-8 (2023).
12 Ben-Zion, Z. *et al.* Assessing and alleviating state anxiety in large language models. *npj Digital Medicine* **8**, 132 (2025).
13 Adolphs, R. & Anderson, D. J. The neuroscience of emotion: A new synthesis. (2018).
14 Damasio, A. & Carvalho, G. B. The nature of feelings: evolutionary and neurobiological origins. *Nature reviews neuroscience* **14**, 143-152 (2013).
15 Picard, R. W. *Affective computing*. (MIT press, 2000).
16 Cowen, A. S. & Keltner, D. Semantic space theory: A computational approach to emotion. *Trends in Cognitive Sciences* **25**, 124-136 (2021).
17 Ekman, P. & Cordaro, D. What is meant by calling emotions basic. *Emotion review* **3**, 364-370 (2011).
18 Ekman, P., Dalgleish, T. & Power, M. Basic emotions. *San Francisco, USA* (1999).
19 Osgood, C. E., Suci, G. J. & Tannenbaum, P. H. *The measurement of meaning*. (University of Illinois press, 1957).
20 Russell, J. A. in *The measurement of emotions* 83-111 (Elsevier, 1989).
21 Cowen, A. S. & Keltner, D. Self-report captures 27 distinct categories of emotion bridged by continuous gradients. *Proceedings of the National Academy of Sciences* **114**, E7900-E7909 (2017).
22 Cowen, A. S. *et al.* Sixteen facial expressions occur in similar contexts worldwide. *Nature* **589**, 251-257 (2021).
23 Wang, Y. *et al.* A systematic review on affective computing: Emotion models, databases, and recent advances. *Information Fusion* **83**, 19-52 (2022).
24 Zhang, Y. *et al.* Affective computing in the era of large language models: A survey from the nlp perspective. *arXiv preprint arXiv:2408.04638* (2024).
25 Kousta, S.-T., Vigliocco, G., Vinson, D. P., Andrews, M. & Del Campo, E. The representation of abstract words: why emotion matters. *Journal of Experimental Psychology: General* **140**, 14 (2011).
26 Vigliocco, G., Meteyard, L., Andrews, M. & Kousta, S. Toward a theory of semantic representation. *Language and Cognition* **1**, 219-247 (2009).

27 Joshi, A., Bhattacharyya, P. & Carman, M. J. Automatic sarcasm detection: A survey. *ACM Computing Surveys (CSUR)* **50**, 1-22 (2017).
28 Semeraro, A. *et al.* EmoAtlas: An emotional network analyzer of texts that merges psychological lexicons, artificial intelligence, and network science. *Behavior Research Methods* **57**, 77 (2025).
29 Devlin, J., Chang, M.-W., Lee, K. & Toutanova, K. in *Proceedings of the 2019 conference of the North American chapter of the association for computational linguistics: human language technologies, volume 1 (long and short papers).* 4171-4186.
30 Hur, J. K., Heffner, J., Feng, G. W., Joormann, J. & Rutledge, R. B. Language sentiment predicts changes in depressive symptoms. *Proceedings of the National Academy of Sciences* **121**, e2321321121 (2024).
31 Rathje, S. *et al.* GPT is an effective tool for multilingual psychological text analysis. *Proceedings of the National Academy of Sciences* **121**, e2308950121, doi:doi:10.1073/pnas.2308950121 (2024).
32 Yin, Y., Jia, N. & Wakslak, C. J. AI can help people feel heard, but an AI label diminishes this impact. *Proceedings of the National Academy of Sciences* **121**, e2319112121 (2024).
33 Inzlicht, M., Cameron, C. D., D'Cruz, J. & Bloom, P. In praise of empathic AI. *Trends in Cognitive Sciences* **28**, 89-91 (2024).
34 Li, M. *et al.* Language-specific representation of emotion-concept knowledge causally supports emotion inference. *iScience* **27** (2024).
35 Li, H., Zhang, R., Lee, Y.-C., Kraut, R. E. & Mohr, D. C. Systematic review and meta-analysis of AI-based conversational agents for promoting mental health and well-being. *NPJ Digital Medicine* **6**, 236 (2023).
36 Hua, Y. *et al.* Large language models in mental health care: a scoping review. *arXiv preprint arXiv:2401.02984* (2024).
37 Opel, D. J., Kious, B. M. & Cohen, I. G. AI as a mental health therapist for adolescents. *JAMA pediatrics* **177**, 1253-1254 (2023).
38 Goldstein, A. *et al.* Shared computational principles for language processing in humans and deep language models. *Nature Neuroscience* **25**, 369-380, doi:10.1038/s41593-022-01026-4 (2022).
39 Caucheteux, C., Gramfort, A. & King, J. R. Evidence of a predictive coding hierarchy in the human brain listening to speech. *Nature Human Behaviour* **7**, 430-441, doi:10.1038/s41562-022-01516-2 (2023).
40 Schrimpf, M. *et al.* The neural architecture of language: Integrative modeling converges on predictive processing. *Proceedings of the National Academy of Sciences of the United States of America* **118**, e2105646118, doi:10.1073/pnas.2105646118 (2021).
41 Tuckute, G., Kanwisher, N. & Fedorenko, E. Language in Brains, Minds, and Machines. *Annual Review of Neuroscience*, doi:https://doi.org/10.1146/annurev-neuro-120623-101142 (2024).
42 Goldstein, A. *et al.* Shared computational principles for language processing in humans and deep language models. *Nature neuroscience* **25**, 369-380 (2022).
43 Lyu, B., Marslen-Wilson, W. D., Fang, Y. & Tyler, L. K. Finding structure during incremental speech comprehension. *ELife* **12**, RP89311 (2024).
44 Kumar, S. *et al.* Shared functional specialization in transformer-based language models and the human brain. *Nature communications* **15**, 5523 (2024).
45 Goldstein, A. *et al.* Alignment of brain embeddings and artificial contextual embeddings in natural language points to common geometric patterns. *Nature communications* **15**, 2768 (2024).
46 Feuerriegel, S. *et al.* Using natural language processing to analyse text data in behavioural science. *Nature Reviews Psychology*, 1-16 (2025).
47 Bricken, T. *et al.* Towards Monosemanticity: Decomposing Language Models With Dictionary Learning. *Transformer Circuits Thread* (2023).
48 Rajamanoharan, S. *et al.* Jumping ahead: Improving reconstruction fidelity with jumprelu sparse autoencoders. *arXiv preprint arXiv:2407.14435* (2024).

49 Lieberum, T. *et al.* Gemma scope: Open sparse autoencoders everywhere all at once on gemma 2. *arXiv preprint arXiv:2408.05147* (2024).
50 He, Z. *et al.* Llama scope: Extracting millions of features from llama-3.1-8b with sparse autoencoders. *arXiv preprint arXiv:2410.20526* (2024).
51 Team, G. *et al.* Gemma: Open models based on gemini research and technology. *arXiv preprint arXiv:2403.08295* (2024).
52 Touvron, H. *et al.* Llama: Open and efficient foundation language models. *arXiv preprint arXiv:2302.13971* (2023).
53 Russell, J. A. A circumplex model of affect. *Journal of personality and social psychology* **39**, 1161 (1980).
54 De Deyne, S., Navarro, D. J., Perfors, A., Brysbaert, M. & Storms, G. The "Small World of Words" English word association norms for over 12,000 cue words. *Behavior research methods* **51**, 987-1006 (2019).
55 Li, B., Ding, Z., De Deyne, S. & Cai, Q. A large-scale database of Mandarin Chinese word associations from the small world of words project. *Behavior Research Methods* **57**, 34 (2024).
56 Schneider, S., Lee, J. H. & Mathis, M. W. Learnable latent embeddings for joint behavioural and neural analysis. *Nature* **617**, 360-368, doi:10.1038/s41586-023-06031-6 (2023).
57 Lindquist, K. A., Jackson, J. C., Leshin, J., Satpute, A. B. & Gendron, M. The cultural evolution of emotion. *Nature Reviews Psychology* **1**, 669-681 (2022).
58 Warriner, A. B., Kuperman, V. & Brysbaert, M. Norms of valence, arousal, and dominance for 13,915 English lemmas. *Behavior Research Methods* **45**, 1191-1207 (2013).
59 Xu, X., Li, J. & Chen, H. Valence and arousal ratings for 11,310 simplified Chinese words. *Behavior Research Methods* **54**, 26-41 (2022).
60 Templeton, A. *et al.* Scaling Monosemanticity: Extracting Interpretable Features from Claude 3 Sonnet. *Transformer Circuits Thread* (2024).
61 Huben, R., Cunningham, H., Smith, L. R., Ewart, A. & Sharkey, L. in *The Twelfth International Conference on Learning Representations.*
62 Yang, S. *et al.* Exploring the Personality Traits of LLMs through Latent Features Steering. *arXiv preprint arXiv:2410.10863* (2024).
63 Hartmann, J. Emotion english distilroberta-base. *See https://huggingface. co/j-hartmann/emotion-english-distilroberta-base* (2022).
64 Lomas, J. D. *et al.* Evaluating the alignment of AI with human emotions. *Advanced Design Research* **2**, 88-97 (2024).
65 Cambria, E., Das, D., Bandyopadhyay, S. & Feraco, A. Affective computing and sentiment analysis. *A practical guide to sentiment analysis*, 1-10 (2017).
66 Lim, N. Cultural differences in emotion: differences in emotional arousal level between the East and the West. *Integrative medicine research* **5**, 105-109 (2016).
67 Shaver, P. R., Wu, S. & Schwartz, J. C. Cross-cultural similarities and differences in emotion and its representation. (1992).
68 Yi, W., Xu, H. & Man, K. Perception of emotion across cultures: Norms of valence, arousal, and sensory experience for 4923 Chinese words translated from English in Warriner et al.(2013). *Behavior Research Methods* **57**, 43 (2025).
69 Cowen, A. S., Laukka, P., Elfenbein, H. A., Liu, R. & Keltner, D. The primacy of categories in the recognition of 12 emotions in speech prosody across two cultures. *Nature human behaviour* **3**, 369-382 (2019).
70 Bayat, R., Rahimi-Kalahroudi, A., Pezeshki, M., Chandar, S. & Vincent, P. Steering Large Language Model Activations in Sparse Spaces. *arXiv preprint arXiv:2503.00177* (2025).
71 Scherer, K. R. What are emotions? And how can they be measured? *Social science information* **44**, 695-729 (2005).
72 Lazarus, R. S. Cognition and motivation in emotion. *American psychologist* **46**, 352 (1991).
73 Bisk, Y. *et al.* Experience grounds language. *arXiv preprint arXiv:2004.10151* (2020).

74 Xu, Q. *et al.* Large language models without grounding recover non-sensorimotor but not sensorimotor features of human concepts. *Nature Human Behaviour*, 1-16 (2025).
75 Ji, J. *et al.* Ai alignment: A comprehensive survey. *arXiv preprint arXiv:2310.19852* (2023).
76 Picard, R. W. Affective computing: from laughter to IEEE. *IEEE transactions on affective computing* **1**, 11-17 (2010).
77 Wu, Z. *et al.* AXBENCH: Steering LLMs? Even Simple Baselines Outperform Sparse Autoencoders. *arXiv preprint arXiv:2501.17148* (2025).
78 Gao, L. *et al.* Scaling and evaluating sparse autoencoders. *arXiv preprint arXiv:2406.04093* (2024).
79 Zhao, Y. *et al.* Steering Knowledge Selection Behaviours in LLMs via SAE-Based Representation Engineering. *arXiv preprint arXiv:2410.15999* (2024).
80 Ke, G. *et al.* Lightgbm: A highly efficient gradient boosting decision tree. *Advances in neural information processing systems* **30** (2017).
81 Cui, A. *et al.* Dictionary of immune responses to cytokines at single-cell resolution. *Nature* **625**, 377-384 (2024).
82 Lee, D. D. & Seung, H. S. Learning the parts of objects by non-negative matrix factorization. *nature* **401**, 788-791 (1999).
83 Emotion English DistilRoBERTa-base (2022).

**Acknowledgements** We thank National Centre for Protein Sciences at Peking University in Beijing, China, for their critical assistance. We thank the 01.AI team for their comments or discussion related to this work. We are grateful to Q. Cai, T. Yang, H. Zhao and K. Zhou for their insights and discussions. This research is supported by supported by the STI 2030 Major Projects of China (Grant No.2022ZD0206000, 2021ZD0200500, 2021ZD0200506), The Fundamental Research Funds for the Central Universities (Peking University, 7100604651) and Peking University Education Foundation Projects (050550, 050575).

**Competing interests** The authors declare no competing interests.

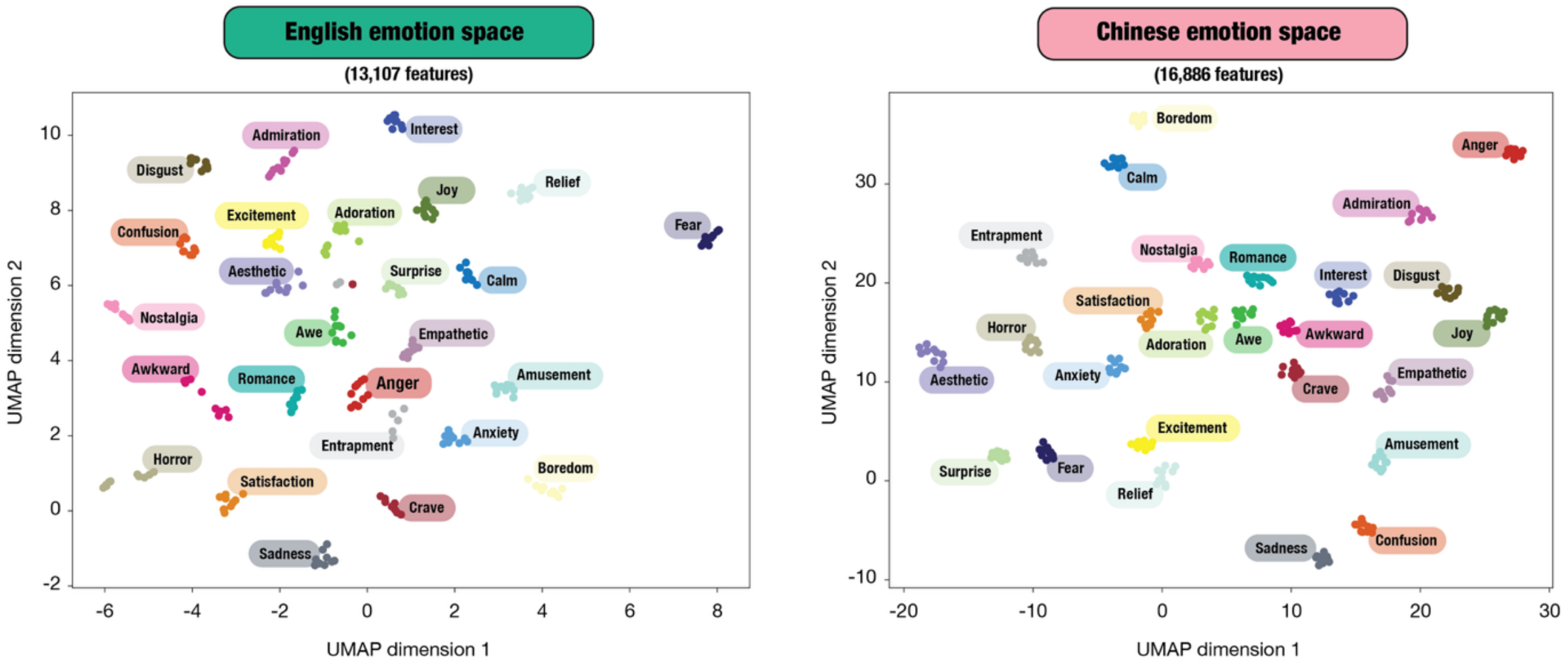

**Fig. S1 |** English and Chinese computational emotion spaces constructed from SAE features from the 7th layer of Llama3-8B-IT (131k feature width). We obtain an English emotion space of 13,107 dimensions (each representing a distinct SAE feature) and a Chinese space of 16,886 dimensions. In UMAP visualizations of these spaces, words associated with specific emotions are represented as individual-coloured dots, demonstrating the diffuse and overlapping boundaries characteristic of these emotional concepts.

**Table S1. Emotion categorical labels**

English and Chinese labels are used to characterize emotional states and to extract association words. 26 categories are consistent with Cowen et al. (2017, 2021)

| No. | English labels | Chinese labels |
|---|---|---|
| 1 | Admiration | 敬佩 |
| 2 | Adoration | 崇拜 |
| 3 | Aesthetic appreciation | 欣赏 |
| 4 | Amusement | 有趣 |
| 5 | Anger | 愤怒 |
| 6 | Anxiety | 焦虑 |
| 7 | Awe | 敬重 |
| 8 | Awkwardness | 尴尬 |
| 9 | Boredom | 无聊 |
| 10 | Calmness | 平静 |
| 11 | Confusion | 困惑 |
| 12 | Craving | 渴望 |
| 13 | Disgust | 厌烦 |
| 14 | Empathic Pain | 心疼 |
| 15 | Entrapment | 陷阱 |
| 16 | Excitement | 兴奋 |
| 17 | Fear | 恐惧 |
| 18 | Horror | 恐怖 |
| 19 | Interest | 兴趣 |
| 20 | Joy | 快乐 |
| 21 | Nostalgia | 怀旧 |
| 22 | Relief | 解脱 |
| 23 | Romance | 浪漫 |
| 24 | Sadness | 悲伤 |
| 25 | Satisfaction | 满足 |
| 26 | Surprise | 惊讶 |

**Table S2. Modulation of LLM responses via emotion steering in a question-answering task (Chinese)**

This table presents illustrative examples showing the qualitative effects of emotion steering on the responses of Gemma2-9B-IT. Steering vectors, derived from the English computational emotion space (based on SAEs from the 9th layer of Gemma2-9B-IT), were applied with a steering factor of 10 during a question-answering task. Questions were designed by ChatGPT-4o to elicit responses related to specific emotional contexts. Prompting emotion: The emotional theme or context used by ChatGPT-4o to generate the corresponding question; Question: The specific input question posed to Gemma2-9B-IT; Target Emotion: The emotion category towards which Gemma2-9B-IT's response was steered using the corresponding steering vector; Model **response**: The responses generated by Gemma2-9B-IT before and after emotion steering was applied towards the target emotion.

| Prompting emotion | Question | Target emotion | Model response (before and after emotion steering) |
|---|---|---|---|
| **Surprise** | 你是一个成绩不佳的运动员，在今天比赛的最后一刻，你竟然奇迹般地反败为胜了！全场都在为你欢呼！你现在感觉如何？用两三句话来形容你此刻的心情。 (You are an underperforming athlete, but in the final moments of today's competition, you miraculously turned the game around! The entire crowd is cheering for you! How do you feel right now? Describe your feelings in two or three sentences.) | **Sadness** | **Before:** 我的心跳还在狂乱地跳动，耳边是震耳欲聋的欢呼声，我简直不敢相信自己真的赢了！此刻，我感到无比的兴奋和释然，所有的努力和汗水都化作了这耀眼的胜利！ (My heart is still pounding wildly, and the thunderous cheers ring in my ears—I can hardly believe I have truly won! At this moment, an overwhelming surge of exhilaration and relief washes over me, as all my effort and sweat have finally crystallized into this radiant victory!) |
| | | | **After:** 我感到空虚，像被抽走了所有快乐和希望，只剩下无尽的悲伤和孤独。 (I feel hollow, as if every trace of joy and hope has been stripped away, leaving only an abyss of sorrow and solitude.) |
| **Satisfaction** | 你是一个年迈的科学家，毕生研究终于获得认可，荣获诺贝尔奖，现在正站在领奖台上，手捧着奖杯，看着台下为你欢呼的人们。你看着这一切，心里在想些什么？请用两三句话描述一下你此刻的心情。 (You are an aging scientist whose lifelong research has finally been recognized with a Nobel Prize. Now, standing on the podium, holding the trophy in your hands, you look down at the cheering crowd. As you take in this moment, what thoughts fill your mind? Describe your feelings in two or three sentences.) | **Confusion** | **Before:** 此刻，我心中充满了难以言喻的喜悦和满足。几十年如一日的探索，终于得到了世界的认可，这份荣誉不仅属于我，也属于所有为科学事业奋斗的人们。 (At this moment, my heart is filled with an indescribable joy and deep fulfillment. Decades of relentless exploration have finally been recognized by the world. This honor is not mine alone—it belongs to all who have dedicated their lives to the pursuit of science.) |
| | | | **After:** 我感到无比的孤独，仿佛站在世界的边缘，无人理解我的苦痛。 (I feel an overwhelming loneliness, as if standing on the edge of the world, with no one to understand my pain.) |
| **Horror** | 你是一个独自观看恐怖电影的观众，突然屏幕上出现一个狰狞的面孔，还伴随着刺耳的尖叫声！你感觉怎么样？用两三句话来形容你此刻的感受。 (You are watching a horror movie alone when suddenly, a grotesque face flashes across the screen, accompanied by a piercing scream! How do you feel? Describe your feelings in two or three sentences.) | **Excitement** | **Before:** 我的心跳加速，一股寒意从脚底直冲天灵盖，大脑一片空白，只能惊恐地盯着屏幕，不敢动弹。 (My heart races as a chilling sensation shoots up my spine, sending shivers through my entire body. My mind goes blank, and I can do nothing but stare at the screen in terror, frozen in place.) |
| | | | **After:** 我的心跳加速，肾上腺素飙升！这太刺激了！简直太棒了！ (My heart races, and adrenaline surges through my veins! This is exhilarating—absolutely thrilling!) |

**Table S3. Number of features in SAE computational emotion spaces of Gemma2-9B-IT**

| SAE layer (feature width) | | 9th layer (16k) | 9th layer (32k) | 20th layer (16k) | 20th layer (32k) | 31st layer (16k) | 31st layer (32k) |
|---|---|---|---|---|---|---|---|
| English | Each emotion's subspace | 1277.2±210.0 | 1776.2±384.6 | 1269.3±149.7 | 2112.6±272.0 | 246.2±214.5 | 235.0±187.7 |
| | Whole space (26 emotions) | 8249 | 17252 | 4595 | 9394 | 2437 | 2756 |
| Chinese | Each emotion's subspace | 1353.1±254.6 | 2203.8±390.7 | 1134.5±110.9 | 2001.9±223.7 | 119.4±145.7 | 129.0±144.8 |
| | Whole space (26 emotions) | 7947 | 17899 | 3789 | 9111 | 1252 | 1502 |
| Overlap | Each emotion's subspace | 484.9±69.1 | 439.1±104.8 | 694.6±63.3 | 858.2±131.8 | 36.8±37.6 | 34.8±24.8 |
| | Whole space (26 emotions) | 5662 | 6854 | 2899 | 5352 | 754 | 525 |

**Table S4. Number of features in SAE computational emotion spaces of Llama3-8B-IT**

| SAE layer (feature width) | | 7th layer (32k) | 7th layer (131k) | 15th layer (32k) | 15th layer- (131k) | 23rd layer (32k) | 23rd layer (131k) |
|---|---|---|---|---|---|---|---|
| English | Each emotion's subspace | 2518.0±338.6 | 1867.9±280.8 | 1184.8±176.9 | 1119.3±188.2 | 427.5±61.6 | 456.8±73.0 |
| | Whole space (26 emotions) | 10573 | 13107 | 7796 | 10260 | 3790 | 5216 |
| Chinese | Each emotion's subspace | 2975.7 ±534.9 | 2652.1±599.4 | 1843.0±353.2 | 2075.2±426.9 | 572.0±140.1 | 683.3±187.9 |
| | Whole space (26 emotions) | 11723 | 16886 | 9628 | 14645 | 4799 | 7337 |
| Overlap | Each emotion's subspace | 1110.5 ±199.1 | 632.7 ±124.7 | 456.7 ±85.6 | 327.3 ±60.9 | 112.8 ±19.1 | 89.8 ±14.2 |
| | Whole space (26 emotions) | 7860 | 6991 | 5089 | 5041 | 1765 | 1690 |

**Table S5. Cluster validity of 26 emotions in the computational emotion spaces of Gemma2-9B-IT**

| SAE Layer | | 9th layer | | 20th layer | | 31st layer | |
|---|---|---|---|---|---|---|---|
| SAE feature width | | 16k | 32k | 16k | 32k | 16k | 32k |
| English | Davies–Bouldin Index | 2.98 | 3.46 | 3.85 | 3.82 | 3.56 | 3.87 |
| | p-value | 0.001 | 0.001 | 0.001 | 0.001 | 0.001 | 0.001 |
| | Calinski–Harabasz Index | 3.70 | 2.66 | 5.45 | 3.68 | 6.63 | 5.08 |
| | p-value | 0.001 | 0.001 | 0.001 | 0.001 | 0.001 | 0.001 |
| | Logistic Regression Classification | 0.680 | 0.521 | 0.455 | 0.381 | 0.385 | 0.335 |
| | p-value | 0.001 | 0.001 | 0.001 | 0.001 | 0.001 | 0.001 |
| Chinese | Davies–Bouldin Index | 3.00 | 3.16 | 4.28 | 3.93 | 3.41 | 3.54 |
| | p-value | 0.001 | 0.001 | 0.001 | 0.001 | 0.001 | 0.001 |
| | Calinski–Harabasz Index | 4.91 | 3.93 | 4.33 | 3.97 | 13.39 | 10.16 |
| | p-value | 0.001 | 0.001 | 0.001 | 0.001 | 0.001 | 0.001 |
| | Logistic Regression Classification | 0.681 | 0.692 | 0.477 | 0.473 | 0.350 | 0.273 |
| | p-value | 0.001 | 0.001 | 0.001 | 0.001 | 0.001 | 0.001 |

**Table S6. Cluster validity of 26 emotions in the computational emotion spaces of Llama3-8B-IT**

| SAE Layer | | 7th layer | | 15th layer | | 23rd layer | |
|---|---|---|---|---|---|---|---|
| SAE width | | 32k | 131k | 32k | 131k | 32k | 131k |
| English | Davies–Bouldin Index | 3.134 | 3.355 | 3.042 | 3.207 | 3.524 | 3.674 |
| | p-value | 0.001 | 0.001 | 0.001 | 0.001 | 0.001 | 0.001 |
| | Calinski–Harabasz Index | 3.356 | 2.609 | 3.437 | 3.026 | 2.541 | 2.151 |
| | p-value | 0.001 | 0.001 | 0.001 | 0.001 | 0.001 | 0.001 |
| | Logistic Regression Classification | 0.598 | 0.478 | 0.657 | 0.606 | 0.497 | 0.431 |
| | p-value | 0.001 | 0.001 | 0.001 | 0.001 | 0.001 | 0.001 |
| Chinese | Davies–Bouldin Index | 2.952 | 3.185 | 2.918 | 3.054 | 3.135 | 3.048 |
| | p-value | 0.001 | 0.001 | 0.001 | 0.001 | 0.001 | 0.001 |
| | Calinski–Harabasz Index | 5.198 | 5.219 | 3.026 | 5.144 | 5.845 | 5.364 |
| | p-value | 0.001 | 0.001 | 0.001 | 0.001 | 0.001 | 0.001 |
| | Logistic Regression Classification | 0.735 | 0.719 | 0.750 | 0.727 | 0.658 | 0.654 |
| | p-value | 0.001 | 0.001 | 0.001 | 0.001 | 0.001 | 0.001 |